%% file: root.tex
\documentclass[letterpaper, 10 pt, conference]{ieeeconf}  

\IEEEoverridecommandlockouts                              

\usepackage{url}
\makeatletter
\let\NAT@parse\undefined
\makeatother
\usepackage[colorlinks, linkcolor=red, anchorcolor=red, citecolor=red]{hyperref}

\usepackage[pdftex]{graphicx}
\usepackage{booktabs, caption, makecell}
\usepackage{threeparttable}
\usepackage{subfig}
\usepackage{balance}
\usepackage[dvipsnames]{xcolor}
\usepackage{multirow}
\usepackage[export]{adjustbox}

\usepackage{amssymb}
\usepackage{pifont}

\usepackage{bbding} 

\newcommand{\cmark}{\ding{51}}%
\newcommand{\xmark}{\ding{55}}%

\title{\LARGE \bf
OdomBeyondVision: An Indoor Multi-modal Multi-platform Odometry Dataset Beyond the Visible Spectrum
}

\author{Peize Li$^{1}$, Kaiwen Cai$^{2}$, Muhamad Risqi U. Saputra$^{3}$, Zhuangzhuang Dai$^{4}$, Chris Xiaoxuan Lu$^{1}$\authorrefmark{1}, \\Andrew Markham$^{4}$ and Niki Trigoni$^{4}$
\thanks{\authorrefmark{1}Corresponding author}
\thanks{This work was partially supported by Amazon Web Services via the Oxford-Singapore Human-Machine Collaboration Programme and EPSRC ACE-OPS (EP/S030832/1)}
\thanks{$^{1}$Peize Li and Chris Xiaoxuan Lu are with the School of Informatics, University of Edinburgh, United Kingdom.
{\tt\footnotesize p.li-36@sms.ed.ac.uk, xiaoxuan.lu@ed.ac.uk}
}
\thanks{$^{2}$Kaiwen Cai is with University of Liverpool, United Kingdom.
{\tt\footnotesize k.cai@liverpool.ac.uk}}
\thanks{$^{3}$Muhamad Risqi U. Saputra is with Monash University, Indonesia.
{\tt\footnotesize risqi.saputra@monash.edu}}
\thanks{$^{4}$Zhuangzhuang Dai, Andrew Markham and Niki Trigoni are with University of Oxford, United Kingdom.
{\tt\footnotesize \{zhudai, andrew.markham, niki.trigoni\}@cs.ox.ac.uk}}
}

\begin{document}

\maketitle
\thispagestyle{empty}
\pagestyle{empty}

\input{sections/0_abstract.tex}

\input{sections/1_introduction.tex}
\input{sections/2_related_work.tex}

\input{sections/3_dataset.tex}
\input{sections/4_benchmarks.tex}
\input{sections/5_conclusion.tex}

\balance
\bibliographystyle{IEEEtran}
\bibliography{IEEEabrv, mybib.bib}

\end{document}

%% file: sections/0_abstract.tex
\begin{abstract}


This paper presents a multimodal indoor odometry dataset, OdomBeyondVision, featuring multiple sensors across the different spectrum and collected with different mobile platforms. Not only does OdomBeyondVision contain the traditional navigation sensors, sensors such as IMUs, mechanical LiDAR, RGBD camera, it also includes several emerging sensors such as the single-chip mmWave radar, LWIR thermal camera and solid-state LiDAR.
With the above sensors on UAV, UGV and handheld platforms, we respectively recorded the multimodal odometry data and their movement trajectories in various indoor scenes and different illumination conditions. We release the exemplar radar, radar-inertial and thermal-inertial odometry implementations to demonstrate their results for future works to compare against and improve upon. The full dataset including toolkit and documentation is publicly available at: \url{https://github.com/MAPS-Lab/OdomBeyondVision}.

\end{abstract}

%% file: sections/1_introduction.tex
\section{Introduction}


Ego-motion estimation, aka. odometry estimation, finds a plethora of applications ranging from the localization and mapping for mobile robots to the pose tracking of wearable devices. By leveraging different navigation sensors (e.g., cameras, LiDAR, and inertial measurement unit (IMU)), last decades witnessed an upsurge of egomotion estimation systems proposed in various use cases \cite{zhang2014loam, campos2021orb, qin2018vins}. These odometry systems, however, all have intrinsic limitations under certain conditions attributed to their underlying sensing mechanism. To name a few, LiDAR data suffers from self-similar structures and environments full of airborne particles (e.g., smoke), while RGB cameras fall short in low illumination and high dynamic range scenarios. IMUs are susceptible to the drifts in large distances or a large magnitude of movement. Realizing the intrinsic limitation of individual sensors, multimodal odometry that jointly uses heterogeneous navigation sensors has received recent attention to enhance the overall robustness of an egomotion estimation. The success of such an odometry system starts from the public availability of multi-sensor datasets which enable systematic testing and community benchmark of different algorithms.

\begin{figure}
    \centering
    \includegraphics[width=3.3in]{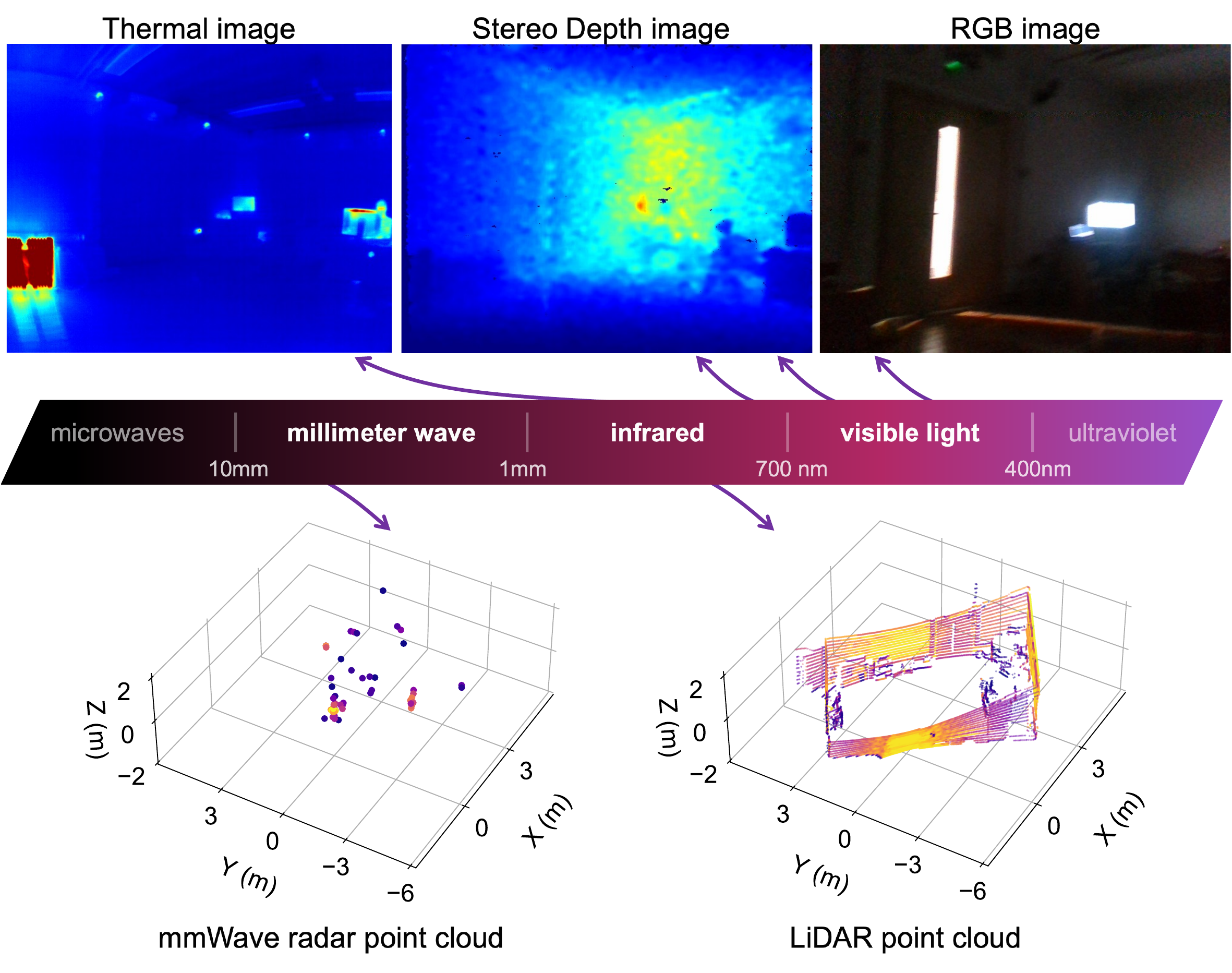}
    \caption{A data example captured in poor illumination from our multimodal odometry dataset. In addition to the above five modalities, our dataset also includes solid-state LiDAR and IMU sensors.}
    \label{openfig}
\end{figure}

\begin{table*}[!t]
\renewcommand{\arraystretch}{1.3}
\caption{Comparison of multimodal odometry datasets}
\label{table_example}
\centering

\begin{threeparttable}
\begin{tabular}{cccccccccc}
\hline
Dataset & Year & Environment & Platform & Radar & LiDAR & RGB & Stereo & Thermal & IMU\\
\hline
KAIST Day/Night\cite{choi2018kaist} & 2018 & Urban & Car & \xmark & \cmark & \cmark & \cmark & \cmark & \cmark\\
Oxford Radar RobotCar\cite{barnes2020oxford} & 2020 & Urban & Car & \cmark & \cmark & \cmark & \cmark & \xmark & \cmark\\
USVInland\cite{cheng2021we} & 2021 & Inland Water & USV & \cmark & \cmark & \cmark & \cmark & \xmark & \cmark \\
ViViD\cite{Lee2019ViViDV} & 2019 & In-/outdoor & Handheld & \xmark & \cmark & \cmark & \cmark & \cmark & \cmark \\
ColoRadar\cite{kramer2021coloradar} & 2020 & In-/outdoor & Handheld/Longboard & \cmark & \cmark & \xmark & \xmark & \xmark & \cmark \\
Weichen Dai et al.\cite{dai2020dataset} & 2020 & In-/outdoor & Handheld & \xmark & \xmark & \cmark & \cmark & \cmark & \cmark\\
Ours & 2022 & Indoor & Handheld/UGV/UAV & \cmark & \cmark & \cmark & \cmark & \cmark & \cmark\\
\hline
\end{tabular}
\end{threeparttable}
\end{table*}

While there exists quite some multimodal odometry datasets \cite{dai2020dataset, Lee2019ViViDV, kim2020mulran, choi2018kaist} in the community, these datasets are limited to the \emph{diversity} in terms of both the navigation sensors onboard and the mobile platforms carrying them. Moreover, multimodal datasets to date are found heavily skewed towards outdoor or urban scenarios, with the indoor datasets mainly dictated by the visual or visual-inertial odometry \cite{schubert2018vidataset, cortes2018advio}. Unfortunately, such a scenario skewness does not coincide with the recent breakthroughs of miniaturised sensors such as single-chip mmWave radars, solid-state sensors and portable thermal cameras. These miniaturized sensors not only become increasingly cost-effective and portable than ever, but also complement the electromagnetic (EM) spectrum covered by conventional navigation sensors. For example, compared with the cameras that only occupy the visible spectrum, mmWave radars and Long Wave Infrared (LWIR) thermal cameras operate at 77~GHz ($4\,mm$ wavelength) and 30~THz ($10\,\mu m$ wavelength) respectively. The extra EM spectrum covered by these emerging sensors can in turn, yield more sensing robustness against poor illumination, airborne particles and visual aliasing etc.

Towards a more \emph{comprehensive} multimodal dataset for odometry evaluation, this work presents a new odometry dataset constituted of the sensory data covering a wide range of electromagnetic frequencies (c.f., Fig.\ref{openfig}). To our best knowledge, this presented dataset features the widest range of mobile platforms and includes robust navigation sensors that emerged recently. Specifically, we recorded 6-Degree-of-Freedom (6DoF) motion with accurate ground truth on  (i) Unmanned Aerial Vehicle (UAV), (ii) Unmanned Ground Vehicle (UGV) and (iii) handheld platforms. We equip these platforms with the emerging mmWave radars, portable thermal cameras and solid-state LiDARs, co-located with the traditional navigation modalities such as spinning LiDAR, RGB-D camera and IMUs. We hope that by making available our data from both emerging and conventional sensors, unprecedented sensor fusion approaches can be yielded on various platforms and different use cases in the wild. 
The main contributions of our datasets are summarised as follows:

\begin{itemize}
    \item We present an indoor odometry dataset containing the data from both emerging and traditional navigation sensors on diverse platforms. This is the first indoor dataset that simultaneously covers the mmWave and LWIR thermal spectrum.
    \item We record the odometry data in 7 different categories of indoor scenes which feature dynamic illumination and various motion patterns. The whole dataset exceeds 10~km in trajectory length.
    \item We provide extrinsic calibration parameters for all platforms and intrinsic parameters for all optic sensors, accompanied by several plug-and-play codes for developing multimodal odometry systems with our dataset.
\end{itemize}

\begin{figure*}
    \centering
    \includegraphics[width=6in]{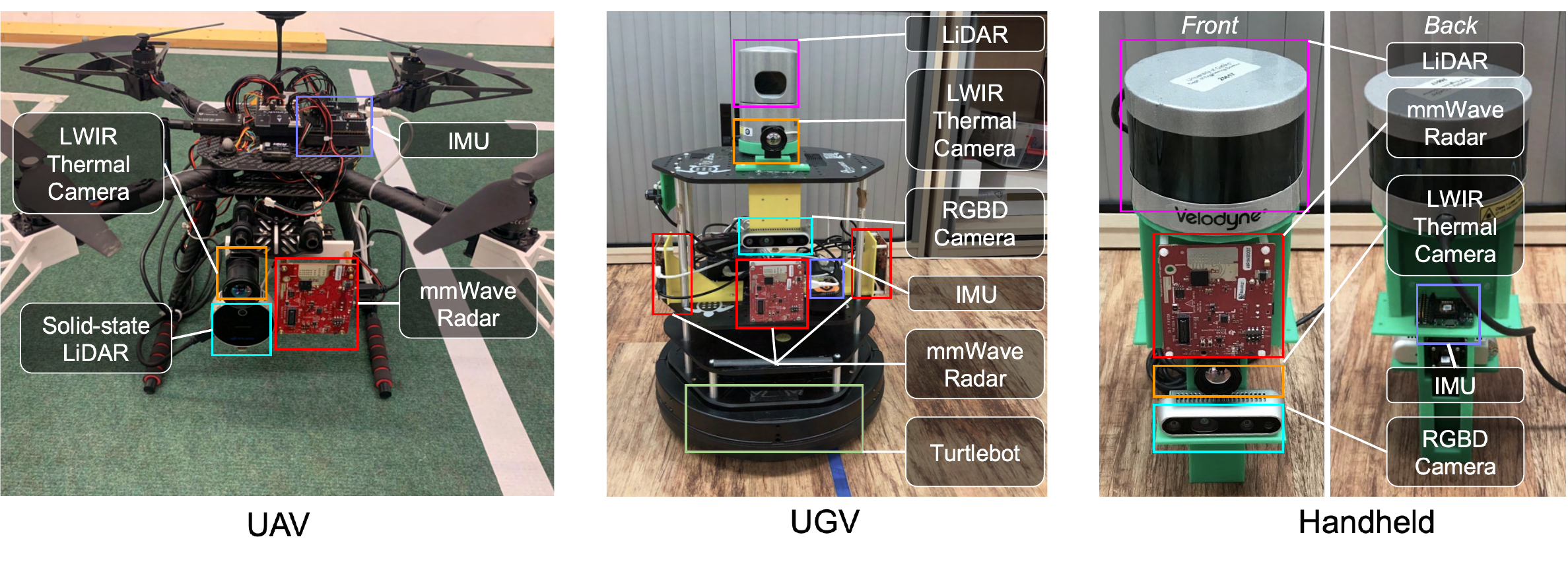}
    \caption{Our sensor setup for UAV, UGV and handheld platform.}
    \label{rig}
\end{figure*}

%% file: sections/2_related_work.tex
\section{Related Work}
A multitude of odometry datasets have been proposed over the last decades, but most of them focus on common visual or LiDAR modalities, while ignoring the emerging radar and thermal sensors. We refer the readers interested in traditional multimodal odometry datasets to \cite{zhan2020visual,geiger2013vision}, and discuss the work related to:

\noindent \textbf{LWIR Thermal Cameras.}
Recent years have seen multiple indoor datasets utilizing the LWIR thermal spectrum. The SubT-Tunnel dataset \cite{rogers2020test} features trajectories in an underground mine from the DARPA Subterranean challenge. Since it is based on the UGV platform only, its movement is restricted to 2D. Indoor trajectories in the ViVid dataset \cite{Lee2019ViViDV} are limited to their Vicon room. This guarantees they have accurate ground truth pose and trajectory, but it also means they do not contain enough scene variety. The multi-spectral dataset \cite{dai2020dataset} is another RGB/thermal camera dataset featuring hardware time synchronization.

\noindent \textbf{mmWave Radars.}
In the meantime, datasets that include radars are mostly recorded for developing advanced driver assistance systems (ADAS). Notable examples include Oxford Radar RobotCar \cite{barnes2020oxford} and nuScenes \cite{caesar2020nuscenes} datasets. However, the radar used in \cite{barnes2020oxford} is a mechanical spinning radar while the single-chip radar used in \cite{caesar2020nuscenes} can only give point clouds in a 2D plane.
The only radar dataset comparable to ours in terms of sensors and scenes is the ColoRadar dataset \cite{kramer2021coloradar}, which features a single-chip mmWave radar, and a cascaded radar. However, the lack of RGB, thermal and Stereo modalities means it cannot support the comparison between radar algorithms and other more popular modalities in the same situation. Furthermore, due to the sparse point cloud generated by radars, our wide-fov 3-radar setup on the UGV platform provides a lot more information than the forward-facing radar setup in ColoRadar.

Unlike all the above works, to our best knowledge, our dataset is the first one that \emph{simultaneously} contain the LWIR camera and mmWave radar sensors, while featuring long trajectories in diversified indoor environments, wide-angle radar arrays, and three different platforms. 

\noindent \textbf{Mobile Platforms.}
Platform-wise, different mobile platforms have different movement characteristics and impose different challenges for odomerty estimation. Compared to other indoor multi-modal datasets that mostly only use handheld platform\cite{Lee2019ViViDV, dai2020dataset}, our dataset is the only one that includes all three viable mobile agents in an indoor environment: UAV, UGV and handheld.

%% file: sections/3_dataset.tex
\section{Dataset}

\subsection{Sensor Setup}

\begin{figure*}[!t]
\centering
\includegraphics[width=1.3in]{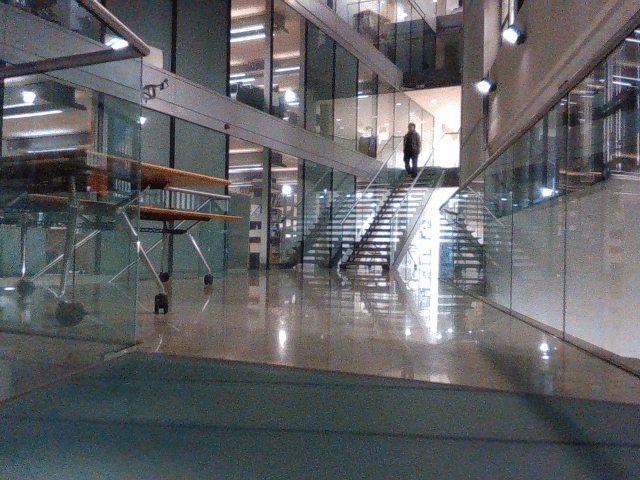}
\hfil
\includegraphics[width=1.3in]{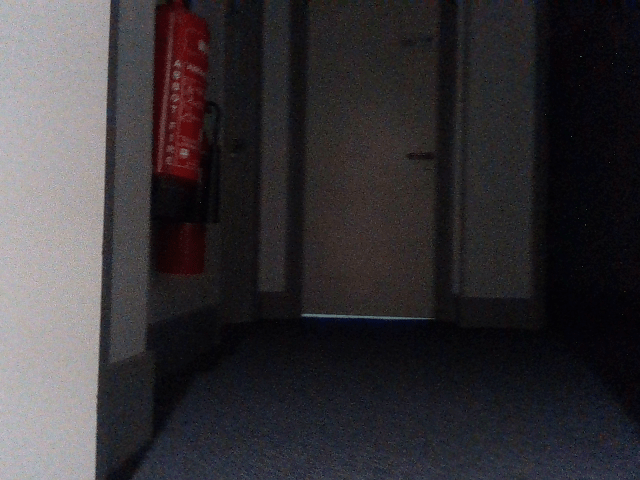}
\hfil
\includegraphics[width=1.3in]{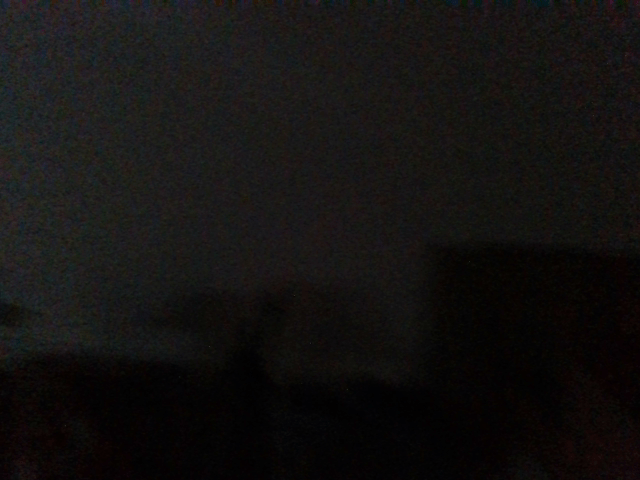}
\hfil
\includegraphics[width=1.3in]{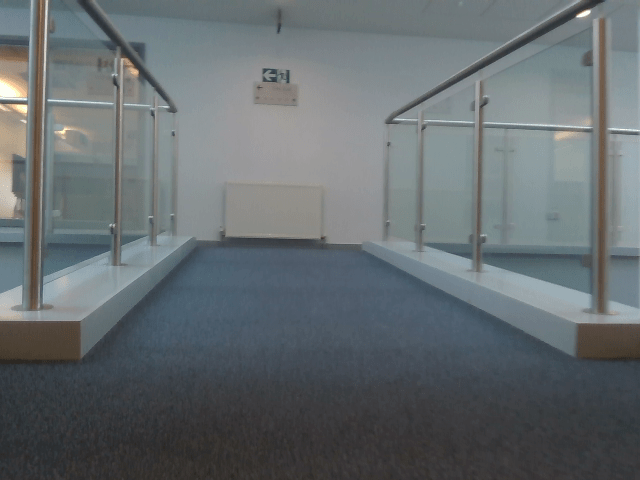}
\hfil
\includegraphics[width=1.3in]{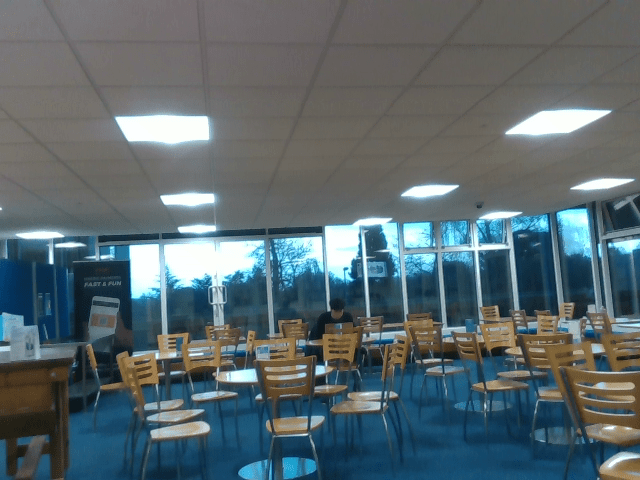}
\\
\subfloat[]{\includegraphics[width=1.3in]{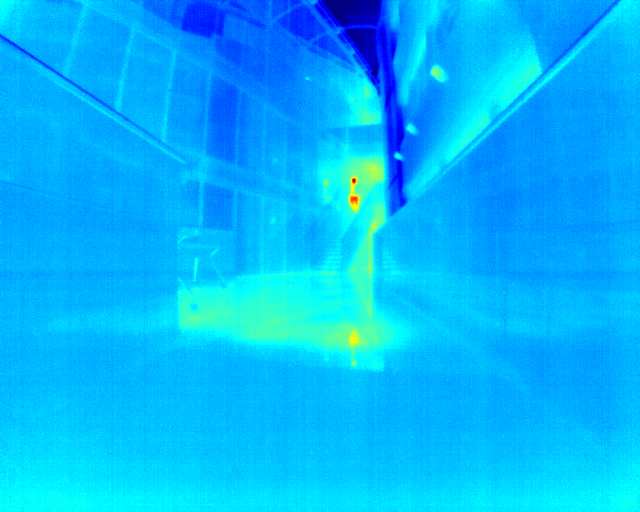}%
\label{scene-1}}
\hfil
\subfloat[]{\includegraphics[width=1.3in]{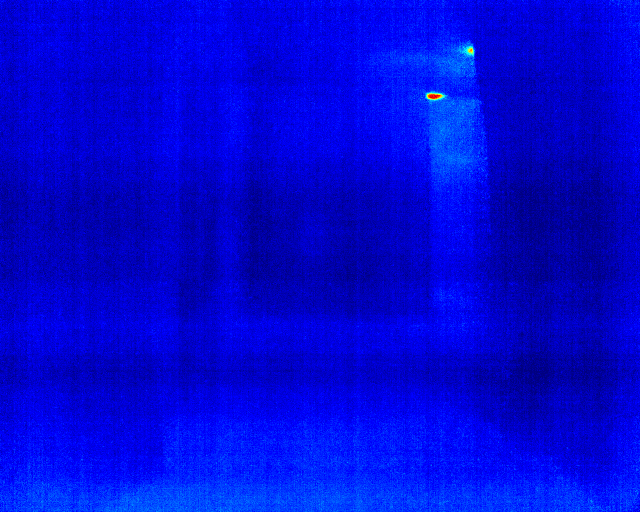}%
\label{scene-2}}
\hfil
\subfloat[]{\includegraphics[width=1.3in]{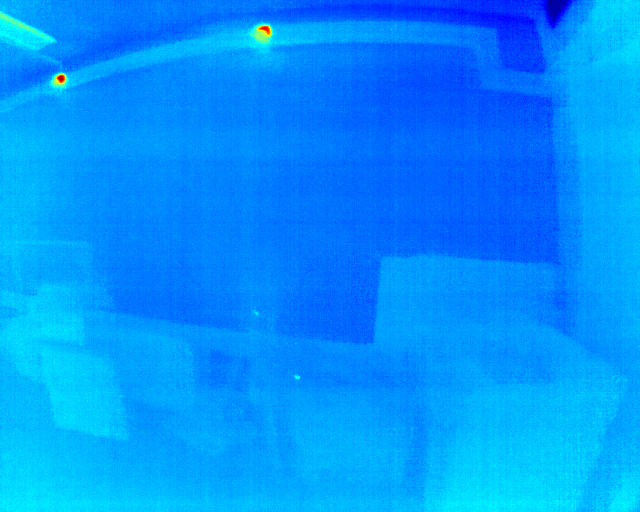}%
\label{scene-3}}
\hfil
\subfloat[]{\includegraphics[width=1.3in]{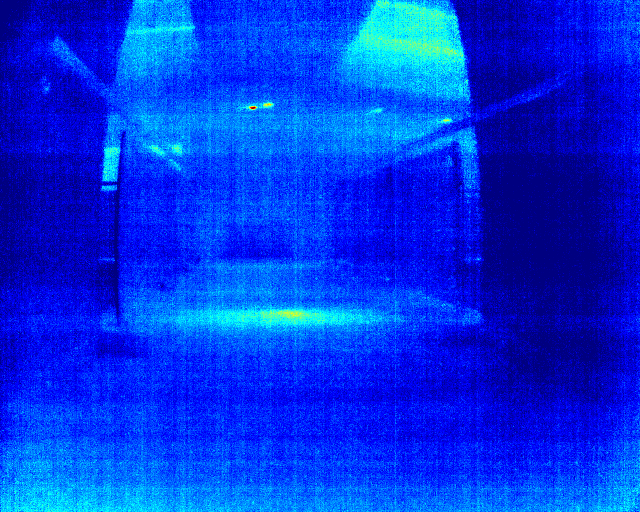}%
\label{scene-4}}
\hfil
\subfloat[]{\includegraphics[width=1.3in]{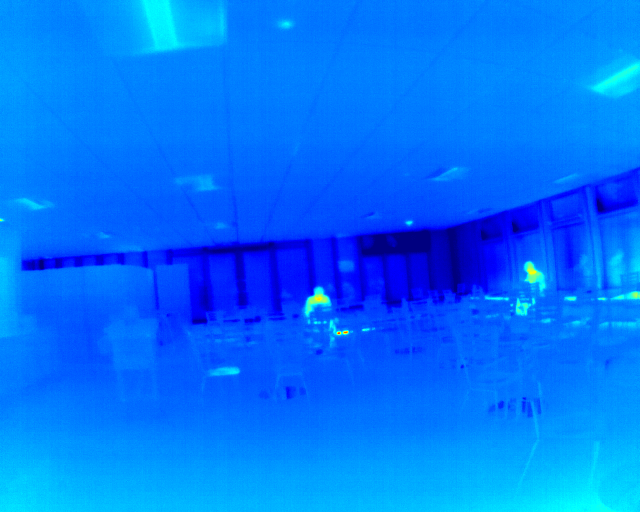}%
\label{scene-5}}
\caption{Examples of Scene Variety - top row from RGB cameras and bottom row from co-located LWIR thermal cameras. We took our sensor rigs to 4 different buildings to collect a variety of scenes. (a) to (c) show different illumination levels, from bright with glares to dimmed to complete darkness. (d) shows large quantities of reflective and transparent surfaces, while (e) shows complex 3D topology in the canteen area.}
\label{scene-variety}
\end{figure*}

\begin{table}
\centering
\caption{Sensor specs. on three platforms.}
\label{table_sensor}
\begin{tabular}{l|l} 
\toprule
\multicolumn{2}{c}{\textit{Handheld}}                                                                                                                         \\ 
\toprule
1x RGBD Camera    & 
\begin{tabular}[c]{@{}l@{}}Intel RealSense D435i stereo camera \\ 20Hz 8bit 640x480 RGB image \\ 30Hz 16bit 640x480 depth image\end{tabular}  \\ 
\hline
1x Thermal Camera & 
\begin{tabular}[c]{@{}l@{}}FLIR Boson 640 thermal camera \\30Hz 16bit 640x512 thermal image\end{tabular}\\ 
\hline
1x LiDAR          & 
\begin{tabular}[c]{@{}l@{}}Velodyne VLP-16 LiDAR \\10Hz 16-channel point cloud\end{tabular}\\ 
\hline
1x Radar          & 
\begin{tabular}[c]{@{}l@{}}TI AWR1843BOOST mmWave radar \\20Hz point cloud\end{tabular}\\
\hline
1x IMU            & 
\begin{tabular}[c]{@{}l@{}}Xsens MTi \\100Hz inertial data and orientation\end{tabular}\\
\bottomrule
\multicolumn{2}{c}{\textit{UGV}}                                                                                                                              \\ 
\toprule
1x RGBD Camera    & 
\begin{tabular}[c]{@{}l@{}}Intel RealSense D435i stereo camera \\ 30Hz 8bit 640x480 RGB image \\ 30Hz 16bit 640x480 depth image\end{tabular}\\
\hline
1x Thermal Camera & 
\begin{tabular}[c]{@{}l@{}}FLIR Boson 640 thermal camera \\ 30Hz 16bit 640x512 thermal image\end{tabular} \\ 
\hline
1x LiDAR          &                                                                         \begin{tabular}[c]{@{}l@{}}Velodyne HDL-32E LiDAR \\ 10Hz 32-channel point cloud\end{tabular}\\ 
\hline
3x Radar          & 
\begin{tabular}[c]{@{}l@{}}TI AWR1843BOOST radars at 90deg angles\\ 20Hz point cloud\end{tabular} \\ 
\hline
1x IMU            & 
\begin{tabular}[c]{@{}l@{}}Xsens MTi \\100Hz inertial data and orientation\end{tabular}\\
\bottomrule
\multicolumn{2}{c}{\textit{UAV}}                                                                                                                              \\ 
\toprule
1x Solid-state LiDAR    & 
\begin{tabular}[c]{@{}l@{}} Intel RealSense L515 Solid-state LiDAR \\15Hz 8bit 960x540 RGB image \\30Hz 16bit 640x480 depth image\end{tabular} \\ 
\hline
1x Thermal Camera & 
\begin{tabular}[c]{@{}l@{}}FLIR Boson 640 thermal camera \\8.7Hz 8bit 640x512 thermal images \end{tabular} \\ 
\hline
1x Radar          & 
\begin{tabular}[c]{@{}l@{}}TI AWR1843BOOST mmWave radar\\ 10Hz point cloud\end{tabular} \\ 
\hline
1x IMU            & 
\begin{tabular}[c]{@{}l@{}}Xsens MTi \\100Hz inertial data and orientation\end{tabular}\\
\bottomrule
\end{tabular}
\end{table}

We collected the data from three platforms, handheld, UAV and UGV, shown in Fig. \ref{rig}. An overview of sensors and their platforms and characteristics are presented in Table. \ref{table_sensor}. The sensor kit includes Intel Realsense D435 stereo camera, RealSense L515 Solid-state LiDAR, FLIR Boson thermal camera, Velodyne VLP-16 and VLP-32, TI AWR1843 mmWave radar and Xsens MTi IMU. UGV has the most room for sensor installment, thus, we used Velodyne HDL-32E and two more radars than on the Handheld platform. On the UAV platform, we switched D435i stereo camera and Velodyne LiDAR for an Intel RealSense L515 Solid-state LiDAR due to the limited payloads of drones.




\subsection{Dataset Collection}

Indoor scenes can vary in many aspects, and such data variety impacts many models, especially learning-based ones, in terms of generalizability and  transferability. In this regard, we collect the dataset under diverse environments and motion patterns. Detailed environment and motion description in each sequence is provided along-side the dataset. We give a brief summary in this section. 

\subsubsection{Sequences} Our dataset collects 44 handheld sequences and 75 UGV sequences through multiple floors in 4 different buildings, providing a large variety of different visual traits, spatial characteristics and illumination conditions. Representative samples are shown in Fig.~\ref{scene-variety}. 3D complexity in our dataset like crowded rooms with densely packed chairs could provide difficulty for range sensors, while darkness, glare, high dynamic range and texture-less walls could be challenging to visual sensors. Several of the sequences include moving targets like humans. This would test a model's robustness to dynamic environments. Due to the need for ground truth from the Vicon tracking system, UAV sequences are recorded in the Vicon room only. All sequences collected from the above platforms are recorded with their 6-DoF movements.

Sequences are named after their characteristics as \emph{[Platform]\_[Scene]\_[TrajectoryShape]\_[LightingCondition]}, examples of which can be seen in Table \ref{ate_table}. For platforms, \emph{H}, \emph{G} and \emph{A} denote handheld, UGV and UAV platforms respectively. For scenes, we divide them into 7 categories:  (1)\emph{vicon} denotes a single room that the platform's motion is confined to, (2)\emph{corridor} means the platform go through narrow long corridors, (3)\emph{glassy} denotes environments with large amounts of reflective surfaces, (4)\emph{atrium} denotes lightly furnished larger spaces, (5)\emph{hallway} denotes large unfurnished spaces, while (6)\emph{canteen} and (7)\emph{office} are heavily furnished spaces. For trajectories, \emph{linear} means long forward-moving trajectories, while \emph{circular} and \emph{eight} denote looped O-shape or 8-shape trajectories with plenty of loop-closure opportunities. For lighting conditions, \emph{bright} denotes normal daylight, \emph{dim} denotes the sequence includes under-exposed RGB frames, and \emph{dark} means the sequence includes pitch-black RGB frames.





\subsubsection{Motion Characteristics} For fully evaluating odometry algorithms, we included a variety of motion patterns when collecting data:

\emph{drastic turns}: The trajectory length ranges from 10 meters to 300 meters. Drastic turns as large as 180 degrees are included in many of the trajectories on all three platforms. Many of the turns make the platform face featureless walls.


\emph{human-controlled operation}: UAV motion is controlled by a human operator with stabilization assistance from the Vicon tracking system. Compared to flight fully controlled by Vicon, control inputs from the human operator are more complex and mimic a genuine trajectory performed by an exploring autonomous agent.

\subsection{Calibration}

To make the dataset applicable to various processing techniques, we provide extrinsic parameters of all sensors and intrinsic parameters of all cameras. Most extrinsic calibrations are done with RGB camera on Realsense D435 as center. A detailed record of coordinate frames and extrinsic parameters for all sensors are provided alongside the dataset.

\subsubsection{Calibrate Thermal Camera vs. RGB Camera} For extrinsic parameters between thermal camera and RGB camera, we need a calibration target that presents good contrast in RGB as well as thermal spectrum. To achieve this, we implemented the masking method of \cite{vidas2012mask}, covering a hot-running laptop with an acrylic mask with hollowed windows.

\begin{figure}[!t]
\centering
\subfloat[Original thermal image]{\includegraphics[width=1.5in]{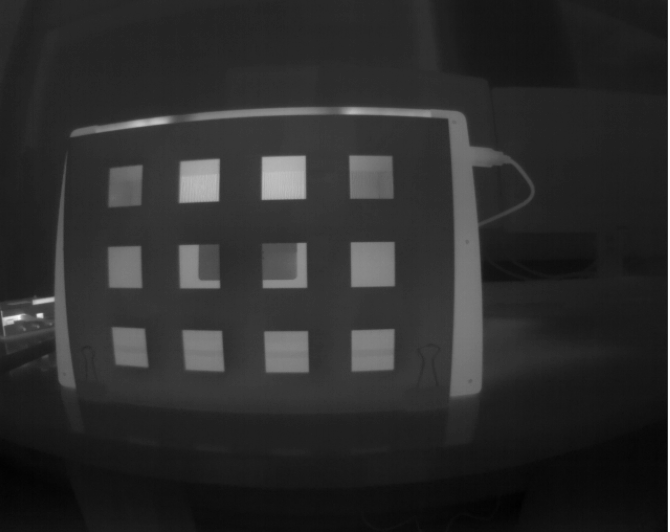}%
\label{thermal-cali-original}}
\hfil
\subfloat[Undistorted thermal image]{\includegraphics[width=1.5in]{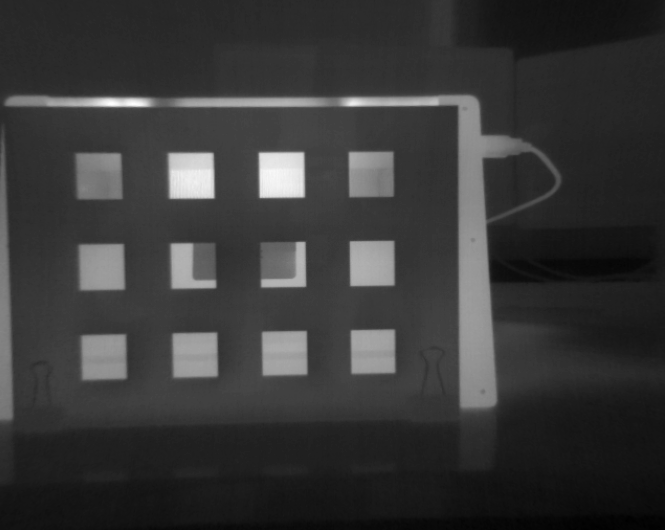}%
\label{thermal-cali-undistort}}
\caption{Thermal masks used to calibrate thermal cameras and their calibration results.}
\label{thermal-cali}
\end{figure}

\subsubsection{Calibrate LiDAR vs. RGB Camera} We implemented the LiDAR vs. RGB camera calibration by Geiger et al.\cite{geiger2012automatic}, using a chessboard patterned corner.

\subsubsection{Calibrate mmWave Radar vs. LiDAR} Due to strong reflection of mmWave signal by metal materials, We use a metallic corner reflector as target to calibrate extrinsic parameters between mmWave radar and LiDAR. Starting from hand measurements, we tune the extrinsic parameter between mmWave radar and LiDAR by finding the best alignment between the corner reflector's signal from the two sensors.


\subsubsection{Calibrate IMU vs. RGB Camera} Extrinsic parameters between IMU and RGB camera are obtained using kalibr tools\cite{rehder2016extending} with AprilTag target.


\subsection{Ground Truth}

For sequences that are collected in large-scale spaces with long trajectories, it's unrealistic to use motion tracking systems (e.g., Vicon) to collect ground truth motions. We follow the practices of \cite{cortes2018advio, doer2021yaw} to use more accurate odometry modalities as pseudo ground truth for modalities with much less accuracy. 

For the UGV platform, we used wheel odometry as the pseudo ground truth. For handheld platform, we use trajectories generated by the SOTA LiDAR odometry, ALOAM \cite{zhang2014loam}, as pseudo ground truth. For the UAV platform, whose trajectories are confined to the Vicon lab, the ground truth comes from the Vicon tracking system, which provides mm-level accuracy 6-DoF poses in 100~Hz. 

\begin{figure*}[!t]
\centering
\includegraphics[width=5.5in]{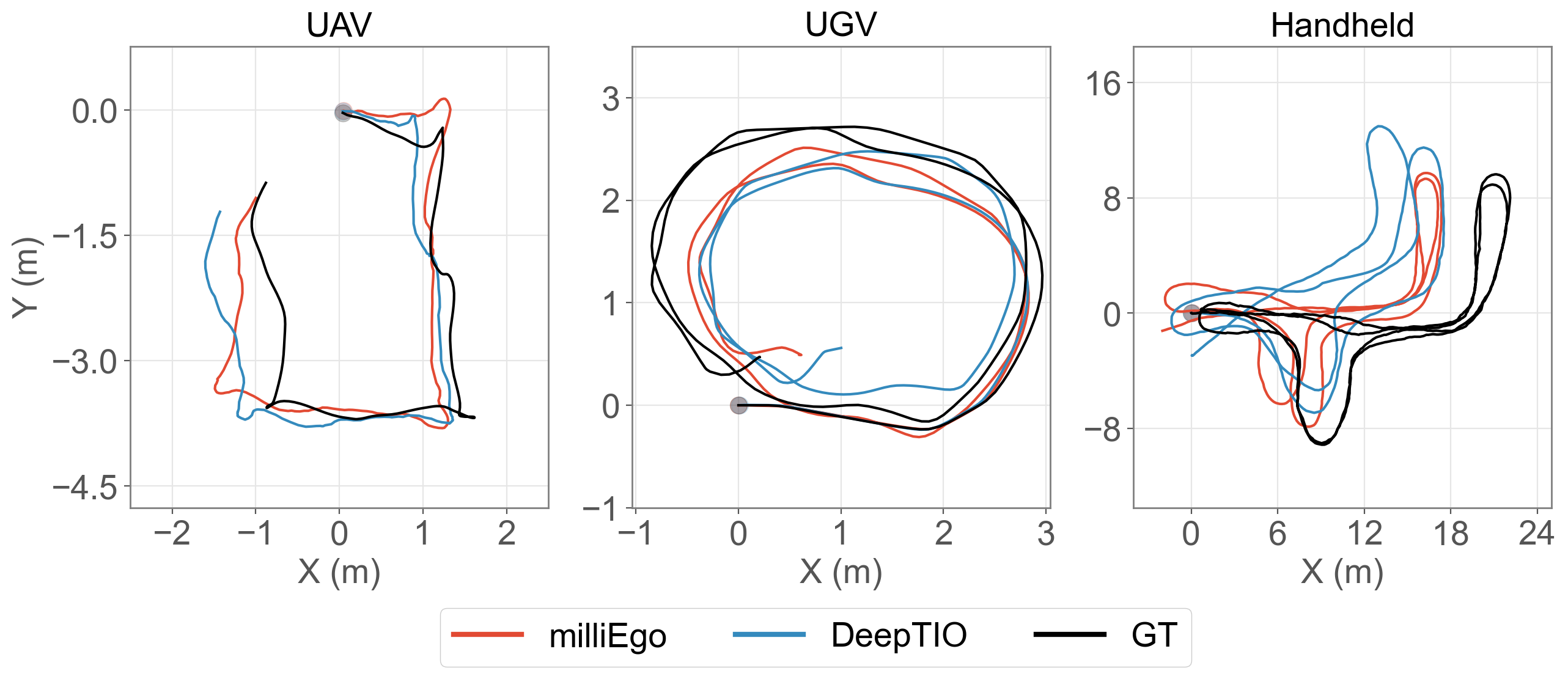}
\caption{Sample Trajectories of TIO and RIO results on UAV, UGV and Handheld platform. Dot denotes the starting point.}
\label{bench}
\end{figure*}

%% file: sections/4_benchmarks.tex
\section{Experiments}


Since the LWIR thermal and mmWave radar spectrum are our particular contribution, we will focus on evaluating our novel thermal and radar data for odometry purposes. Since there are few open-source models involving the fusion of more than one spectrum, we will be implementing radar odometry (RO), radar-inertial odometry (RIO) and thermal-inertial odometry (TIO) methods to estimate 6-DoF trajectories. To demonstrate the dataset in different use-cases, odometry using radar and thermal will each be implemented using a learning-based method and a none-learning based method.

We use two common metrics, Absolute Trajectory Error (ATE)\cite{geiger2012we} and Relative Pose Error (RPE)  \cite{6385773}, to evaluate 6-DoF odometry on our dataset. We perform a train/test split on the dataset that ensures the scenes are evenly split. Along with our dataset, we provide data pre-processing, model training, and model testing codes.

\subsection{Odometry with mmWave Radar}

MilliEgo \cite{lu2020milliego} is a first-of-its-kind deep learning framework for radar-inertial odometry. It uses Convolutional Neural Network (CNN) and Recurrent Neural Network (RNN) to encode radar and IMU streams respectively, after which the 2 encoded streams are fused together using mixed-attention mechanism. The fused sensor features are temporally processed by long short-term memory (LSTM) and regressed by fully-connected (FC) layers.

We trained the network using its open-sourced configuration on our handheld, UGV and UAV platforms, generating pose predictions for each platform at 4~Hz (UGV), 5~Hz (UAV) and 6.7~Hz (handheld). The results shown in Fig. \ref{bench} and Table \ref{ate_table} demonstrate that milliEgo is able to generate good trajectories, with acceptable drift even in long trajectories. As expected, milliEgo performs better on the UGV platform than handheld and UAV platforms with the wide-fov 3-radar array on the UGV compared to only 1 radar on the other platforms.

For the non-learning based method, we implemented a Radar-ICP model, which uses iterative closest point (ICP)\cite{rusinkiewicz2001efficient} to register radar frames at fixed intervals. Adjacent point cloud frames are overlaid to reduce noise. As can be seen in Table \ref{ate_table}, the trajectories proved difficult for Radar-ICP to estimate. The high noise level in the mmWave radar's data can explain the large error in Radar-ICP, which fails to match point cloud frames with many points flickering on and off. Like milliEgo, Radar-ICP also shows better results with 3-radar setup than the 1-radar handheld and UAV platform.

\begin{table*}[!t]
\centering
\captionsetup{justification=centering,margin=2cm}
\caption{Performance of Different Methods on Test Sequences: Mean Absolute Trajectory Errors (ATE) [m] and Mean Relative Pose Error (RPE). RPE shown below is in the format of Translation[m]/Rotation[deg].}
\label{ate_table}
\begin{tabular}{lc|cc|cc|cc}
\toprule
                                    &          & \multicolumn{2}{c}{milliEgo\cite{lu2020milliego}} & \multicolumn{2}{c}{DeepTIO\cite{saputra2020deeptio}} & \multicolumn{2}{c}{ICP\cite{rusinkiewicz2001efficient}} \\
Test Sequence                            & Length[m]/Duration[s] & ATE & RPE & ATE & RPE & ATE & RPE\\
\midrule
\multicolumn{2}{l}{Handheld} & \\
\midrule
H\_corridor\_circle\_bright\_5      & 141.5/162.7 & 8.72  & \textbf{0.071/2.44} &  \textbf{8.29} & 0.087/4.18  & 15.82  & 0.338/13.56 \\
H\_vicon\_circle\_dark\_3           & 58.3/87.0 & 2.49  & \textbf{0.056/3.20} &  \textbf{2.35} & 0.067/5.26  &  5.61  & 0.296/13.03 \\
H\_vicon\_circle\_dark\_4           & 43.1/61.3 & \textbf{1.24}  & \textbf{0.064/2.88} &  5.55 & 0.072/5.13  &  3.14  & 0.278/12.69 \\
H\_viconcorridor\_circle\_bright\_3 & 104.2/114.3 & \textbf{2.70}  & \textbf{0.062/2.68} &  3.36 & 0.076/4.51  &  12.45 & 0.271/12.73 \\
H\_office\_circle\_bright\_2        & 76.0/94.8 & \textbf{2.29}  & \textbf{0.061/3.77} &  5.35 & 0.086/5.83  & 7.33   & 0.345/17.06 \\
H\_atrium\_linear\_bright\_3        & 146.6/122.6 & \textbf{4.37}  & \textbf{0.078/2.95} &  5.54 & 0.101/4.58  & 13.09  & 0.388/11.11 \\
H\_multifloor\_linear\_1            & 318.0/261.8 & \textbf{7.91}  & \textbf{0.096/3.46} &  22.54 & 0.114/5.04 & 18.98  & 0.368/11.24 \\
H\_hallway\_circle\_bright\_2       & 202.2/282.9 & 16.00 & \textbf{0.071/4.03} &  \textbf{13.39} & 0.083/6.26 & 24.81  & 0.317/15.49 \\
H\_glassy\_circle\_bright\_2        & 73.6/86.1 & 7.18  & \textbf{0.077/2.33} &  \textbf{3.58} & 0.082/3.52  & 8.27   & 0.314/9.96 \\
H\_canteen\_circle\_bright\_1       & 153.4/159.3 & \textbf{4.98}  & \textbf{0.080/3.88} &  5.34 & 0.090/5.55  & 10.32  & 0.395/15.77 \\
H\_atrium\_circle\_bright\_1        & 146.8/147.8 & 7.47  & \textbf{0.074/3.05} &  \textbf{7.24} & 0.084/4.68  & 7.88   & 0.400/15.03 \\
\midrule
\multicolumn{2}{l}{UGV} & \\
\midrule
G\_vicon\_eight\_bright\_1          & 24.6/63.2 & 0.54  & 0.022/2.99 &  \textbf{0.46} & \textbf{0.014/1.77} &  1.40  & 0.119/4.83 \\
G\_stairwell\_linear\_bright\_1     & 62.5/139.6 & 1.90  & 0.017/1.50 &  \textbf{1.24} & \textbf{0.014/0.87} &  7.71  & 0.091/3.18 \\
G\_atrium\_linear\_bright\_2        & 91.6/209.1 & \textbf{2.62}  & 0.018/1.82 &  3.23 & \textbf{0.015/0.97} & 29.68  & 0.100/3.73 \\
G\_canteen\_circle\_bright\_1       & 71.0/171.7 & \textbf{1.95}  & 0.019/2.19 &  2.86 & \textbf{0.017/1.27} &  5.58  & 0.102/4.16 \\
G\_hallway\_linear\_bright\_1       & 78.7/183.3 & \textbf{2.19}  & 0.019/1.58 &  4.20 & \textbf{0.015/0.95} & 10.01  & 0.096/3.75 \\
G\_hallway\_linear\_bright\_3       & 66.0/148.87 & \textbf{2.07}  & 0.019/1.67 &  3.98 & \textbf{0.014/0.95} &  6.93  & 0.101/3.72 \\
G\_corridor\_linear\_bright\_11     & 40.2/93.0 & 2.38  & 0.017/1.83 &  \textbf{1.50} & \textbf{0.016/1.06} &  9.39  & 0.111/3.88 \\
G\_corridor\_linear\_bright\_13     & 25.1/57.2 & \textbf{1.32}  & 0.018/1.32 &  1.35 & \textbf{0.013/0.78} &  3.44  & 0.092/3.55 \\
G\_vicon\_eight\_bright\_5          & 61.3/150.0 & 0.80  & 0.019/2.44 &  \textbf{0.56} & \textbf{0.015/1.35} &  2.26 & 0.112/6.02 \\
G\_corridor\_linear\_bright\_18     & 39.3/88.3 & \textbf{0.77}  & 0.018/1.00 &  2.28 & \textbf{0.015/0.56} &  3.11  & 0.098/3.33 \\
\midrule
\multicolumn{2}{l}{UAV} & \\
\midrule
A\_vicon2\_circle\_bright\_6        & 5.55/40.0 & 1.27  & 0.023/\textbf{0.75} &  \textbf{0.62} & \textbf{0.020}/0.96 & 2.51  & 0.451/13.30 \\
A\_vicon2\_circle\_bright\_7        & 11.27/40.0 & 0.45  & 0.024/\textbf{1.41} &  \textbf{0.34} & \textbf{0.022}/1.78 & 3.68  & 0.508/18.54 \\
A\_vicon2\_circle\_bright\_8        & 10.75/40.0 & 1.33  & 0.032/\textbf{1.47} &  \textbf{0.45} & \textbf{0.028}/1.89 & 6.49  & 0.494/16.75 \\
A\_vicon2\_circle\_bright\_9        & 10.98/40.0 & 1.13  & 0.027/\textbf{1.11} &  \textbf{0.55} & \textbf{0.025}/1.52 & 3.32  & 0.510/15.85 \\
A\_vicon2\_circle\_bright\_10       & 11.14/40.0 & 0.76  & 0.025/\textbf{1.20} &  \textbf{0.44} & \textbf{0.023}/1.62 & 5.72  & 0.737/22.45 \\
\bottomrule
\end{tabular}
\end{table*}

\subsection{Odometry with Thermal Camera}

\begin{figure}[!t]
\centering
\subfloat[]{\includegraphics[width=1.05in,valign=t]{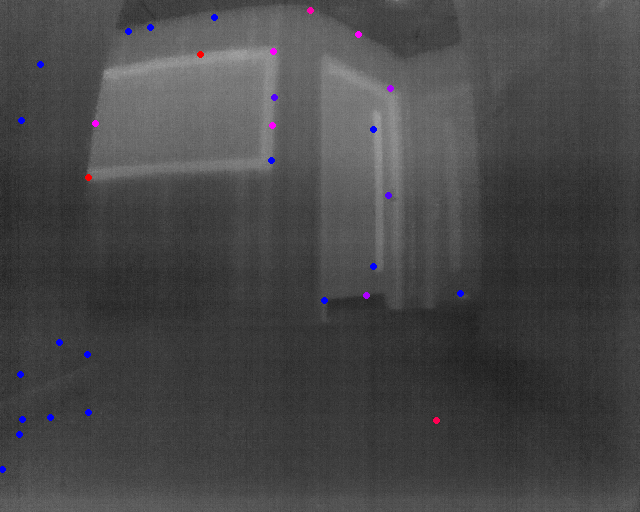}%
\label{vins-before}}
\hfil
\subfloat[]{\includegraphics[width=1.05in,valign=t]{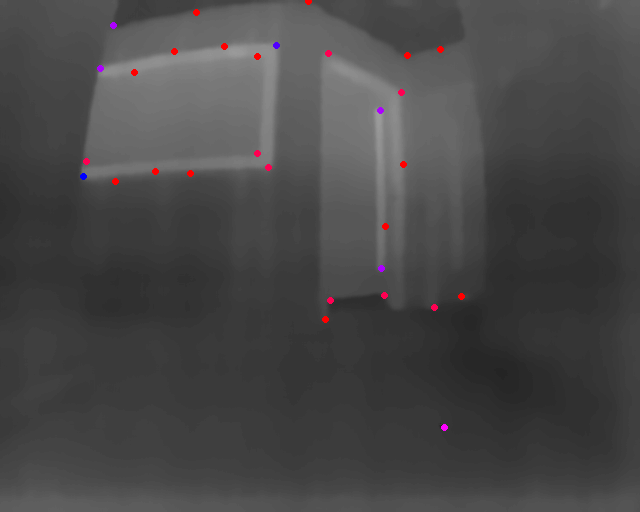}%
\label{vins-after}}
\hfil
\subfloat{\includegraphics[width=1.2in,valign=t]{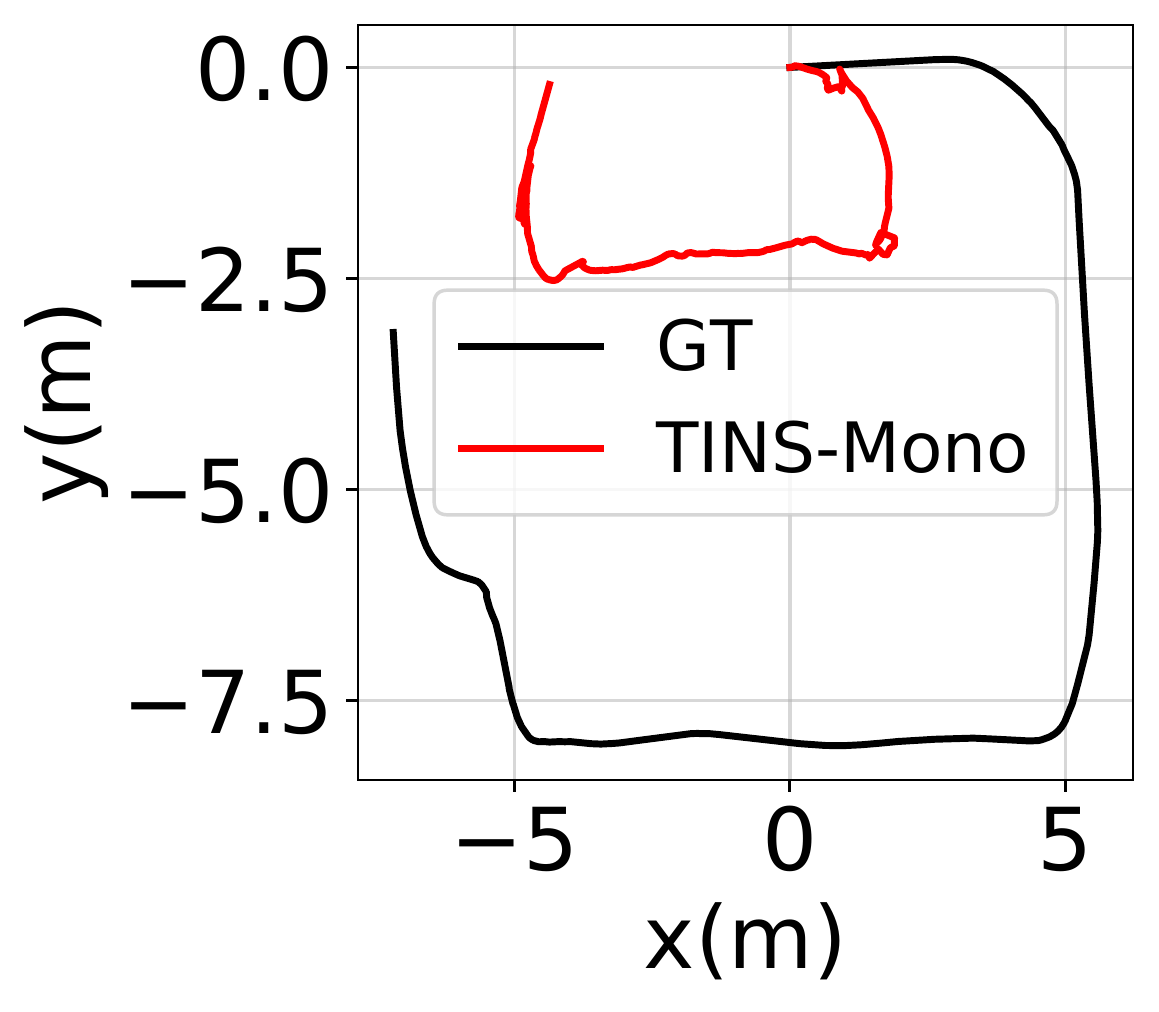}%
\label{vins-track}}
\caption{TINS-Mono tracking and trajectory. (a) and (b) shows TINS-Mono feature tracking before and after our pre-processing. Red dots are features successfully matched with previous frames, while blue dots are failed matches. Noise artifacts are falsely extracted as features in (a). The trajectory shows TINS-Mono severely underestimates translation motion.}
\label{vins}
\end{figure}

Existing Visual-Inertial Odometry (VIO) models can be converted to a TIO since they share the same input format. 
Thus, for our non-learning TIO model, we adapted the popular visual-inertial state estimation framework VINS-Mono \cite{qin2018vins} to TINS-Mono by taking the thermal image as input. Different from the random noise pattern of RGB sensors, thermal sensors accumulate noise on fixed pixel locations (fixed-pattern noise) for a short duration, in which those noises will be typically suppressed through the camera's Non-Uniformity Correction (NUC), but could still be severe enough to affect models.
\cite{saputra2021graph}. Feature extraction techniques used by TINS-Mono could mistake those fixed noise artifacts as visual features. To mitigate this delusion, TINS-Mono requires the thermal images to go through a series of pre-processing including contrast adjustments, denoising and sharpening. As shown in Fig. \ref{vins}, the TINS-Mono is able to stably track features in most scenarios after pre-processing, but the shortage and inconsistency in visual features makes the model underestimate translation motion, and it would still understandably lose track in the presence of thermally-flat images.

DeepTIO \cite{saputra2020deeptio} is a deep learning thermal-inertial odometry method aimed at alleviating the lack of visual features in thermal images. It benefits from a hallucination network that aims to hallucinate upon the thermal images such that it can produce similar visual features as the ones extracted from RGB cameras. In a word, it is equivalent to fusing together 3 modalities: (hallucinated) visual, thermal, and IMU.

We trained the network on handheld, UAV and UGV platform with exactly the same configuration, giving out pose predictions at a rate of 6~Hz (on handheld and UGV) and 4.3~Hz (on UAV). Before training, a pre-trained VINet \cite{clark2017vinet} was used to first generate the visual features from the RGB stream. In the first stage of training, the hallucination network was trained to generate similar features from thermal stream. In the second stage, the rest of the network was trained while the hallucination layers were frozen. The estimated trajectories for DeepTIO are shown in Fig.\ref{bench} and the quantitative results are presented in Table. \ref{ate_table}. Notably, DeepTIO doesn't show noticeable drop in performance in pitch-dark sequences where Visual-Inertial models would completely fail. The better performance of DeepTIO on the UGV than on the handheld may be a result of the unstable motion and faster speed of the handheld devices, making feature tracking slightly more challenging. A customized and tailored network configuration might be required to generate a better trajectory in this challenging scenario.



To summarize our experiments, the evaluated methods show that sensible odometry results can be generated from our dataset using LWIR and mmWave spectrum, but their performance still leaves much to be desired on our challenging dataset. With the two non-learning methods, Radar-ICP and TINS-Mono, we demonstrated the common challenges of the LWIR thermal and mmWave modalities: the high level of noise and the sparsity of features. Comparing results of non-learning and learning based methods, it appears learning based methods consistently outperform non-learning based one across three platforms, while also avoiding manual tuning of model parameters. Platform-wise, handheld and UAV platforms prove to be more challenging than UGV due to their complex 6-DoF movements. We hope that our diverse and challenging dataset would facilitate multi-modal and multi-platform odometry research in the future.



%% file: sections/5_conclusion.tex
\section{Conclusion}
This paper presents the OdomBeyondVision dataset. By including the traditional navigation sensors and the emerging single-chip mmWave radar and LWIR cameras, our odometry dataset features a wide sensor coverage across both visible and invisible spectrums. 
We collected the odometry data with multiple different mobile platforms in a variety of indoor scenes and illumination, aiming to bridge existing gaps and further diversify odometry datasets in the community. 
We demonstrated the use of the dataset in exemplar odometry systems and released their implementations for plug-and-play usage. Owing to the increasing interest in localization and navigation research, our plan in the near future is to provide better dataset services with more benchmarking results.

%% file: root.bbl
\begin{thebibliography}{10}
\providecommand{\url}[1]{#1}
\csname url@samestyle\endcsname
\providecommand{\newblock}{\relax}
\providecommand{\bibinfo}[2]{#2}
\providecommand{\BIBentrySTDinterwordspacing}{\spaceskip=0pt\relax}
\providecommand{\BIBentryALTinterwordstretchfactor}{4}
\providecommand{\BIBentryALTinterwordspacing}{\spaceskip=\fontdimen2\font plus
\BIBentryALTinterwordstretchfactor\fontdimen3\font minus
  \fontdimen4\font\relax}
\providecommand{\BIBforeignlanguage}[2]{{%
\expandafter\ifx\csname l@#1\endcsname\relax
\typeout{** WARNING: IEEEtran.bst: No hyphenation pattern has been}%
\typeout{** loaded for the language `#1'. Using the pattern for}%
\typeout{** the default language instead.}%
\else
\language=\csname l@#1\endcsname
\fi
#2}}
\providecommand{\BIBdecl}{\relax}
\BIBdecl

\bibitem{zhang2014loam}
J.~Zhang and S.~Singh, ``Loam: Lidar odometry and mapping in real-time.'' in
  \emph{Robotics: Science and Systems}, vol.~2, no.~9, 2014.

\bibitem{campos2021orb}
C.~Campos, R.~Elvira, J.~J.~G. Rodr{\'\i}guez, J.~M. Montiel, and J.~D.
  Tard{\'o}s, ``Orb-slam3: An accurate open-source library for visual,
  visual--inertial, and multimap slam,'' \emph{IEEE Transactions on Robotics},
  2021.

\bibitem{qin2018vins}
T.~Qin, P.~Li, and S.~Shen, ``Vins-mono: A robust and versatile monocular
  visual-inertial state estimator,'' \emph{IEEE Transactions on Robotics},
  vol.~34, no.~4, pp. 1004--1020, 2018.

\bibitem{choi2018kaist}
Y.~Choi, N.~Kim, S.~Hwang, K.~Park, J.~S. Yoon, K.~An, and I.~S. Kweon, ``Kaist
  multi-spectral day/night data set for autonomous and assisted driving,''
  \emph{IEEE Transactions on Intelligent Transportation Systems}, vol.~19,
  no.~3, pp. 934--948, 2018.

\bibitem{barnes2020oxford}
D.~Barnes, M.~Gadd, P.~Murcutt, P.~Newman, and I.~Posner, ``The oxford radar
  robotcar dataset: A radar extension to the oxford robotcar dataset,'' in
  \emph{2020 IEEE International Conference on Robotics and Automation
  (ICRA)}.\hskip 1em plus 0.5em minus 0.4em\relax IEEE, 2020, pp. 6433--6438.

\bibitem{cheng2021we}
Y.~Cheng, M.~Jiang, J.~Zhu, and Y.~Liu, ``Are we ready for unmanned surface
  vehicles in inland waterways? the usvinland multisensor dataset and
  benchmark,'' \emph{IEEE Robotics and Automation Letters}, vol.~6, no.~2, pp.
  3964--3970, 2021.

\bibitem{Lee2019ViViDV}
A.~J. Lee, Y.~Cho, S.~Yoon, Y.-S. Shin, and A.~Kim, ``Vivid : Vision for
  visibility dataset,'' 2019.

\bibitem{kramer2021coloradar}
A.~Kramer, K.~Harlow, C.~Williams, and C.~Heckman, ``Coloradar: The direct 3d
  millimeter wave radar dataset,'' \emph{arXiv preprint arXiv:2103.04510},
  2021.

\bibitem{dai2020dataset}
W.~Dai, Y.~Zhang, S.~Chen, D.~Sun, and D.~Kong, ``A dataset for evaluating
  multi-spectral motion estimation methods,'' \emph{arXiv preprint
  arXiv:2007.00622}, 2020.

\bibitem{kim2020mulran}
G.~Kim, Y.~S. Park, Y.~Cho, J.~Jeong, and A.~Kim, ``Mulran: Multimodal range
  dataset for urban place recognition,'' in \emph{2020 IEEE International
  Conference on Robotics and Automation (ICRA)}.\hskip 1em plus 0.5em minus
  0.4em\relax IEEE, 2020, pp. 6246--6253.

\bibitem{schubert2018vidataset}
D.~Schubert, T.~Goll, N.~Demmel, V.~Usenko, J.~Stueckler, and D.~Cremers, ``The
  tum vi benchmark for evaluating visual-inertial odometry,'' in
  \emph{International Conference on Intelligent Robots and Systems (IROS)},
  October 2018.

\bibitem{cortes2018advio}
S.~Cort{\'e}s, A.~Solin, E.~Rahtu, and J.~Kannala, ``Advio: An authentic
  dataset for visual-inertial odometry,'' in \emph{Proceedings of the European
  Conference on Computer Vision (ECCV)}, 2018, pp. 419--434.

\bibitem{zhan2020visual}
H.~Zhan, C.~S. Weerasekera, J.-W. Bian, and I.~Reid, ``Visual odometry
  revisited: What should be learnt?'' in \emph{2020 IEEE International
  Conference on Robotics and Automation (ICRA)}.\hskip 1em plus 0.5em minus
  0.4em\relax IEEE, 2020, pp. 4203--4210.

\bibitem{geiger2013vision}
A.~Geiger, P.~Lenz, C.~Stiller, and R.~Urtasun, ``Vision meets robotics: The
  kitti dataset,'' \emph{The International Journal of Robotics Research},
  vol.~32, no.~11, pp. 1231--1237, 2013.

\bibitem{rogers2020test}
J.~G. Rogers, J.~M. Gregory, J.~Fink, and E.~Stump, ``Test your slam! the
  subt-tunnel dataset and metric for mapping,'' in \emph{2020 IEEE
  International Conference on Robotics and Automation (ICRA)}.\hskip 1em plus
  0.5em minus 0.4em\relax IEEE, 2020, pp. 955--961.

\bibitem{caesar2020nuscenes}
H.~Caesar, V.~Bankiti, A.~H. Lang, S.~Vora, V.~E. Liong, Q.~Xu, A.~Krishnan,
  Y.~Pan, G.~Baldan, and O.~Beijbom, ``nuscenes: A multimodal dataset for
  autonomous driving,'' in \emph{Proceedings of the IEEE/CVF conference on
  computer vision and pattern recognition}, 2020, pp. 11\,621--11\,631.

\bibitem{vidas2012mask}
S.~Vidas, R.~Lakemond, S.~Denman, C.~Fookes, S.~Sridharan, and T.~Wark, ``A
  mask-based approach for the geometric calibration of thermal-infrared
  cameras,'' \emph{IEEE Transactions on Instrumentation and Measurement},
  vol.~61, no.~6, pp. 1625--1635, 2012.

\bibitem{geiger2012automatic}
A.~Geiger, F.~Moosmann, {\"O}.~Car, and B.~Schuster, ``Automatic camera and
  range sensor calibration using a single shot,'' in \emph{2012 IEEE
  international conference on robotics and automation}.\hskip 1em plus 0.5em
  minus 0.4em\relax IEEE, 2012, pp. 3936--3943.

\bibitem{rehder2016extending}
J.~Rehder, J.~Nikolic, T.~Schneider, T.~Hinzmann, and R.~Siegwart, ``Extending
  kalibr: Calibrating the extrinsics of multiple imus and of individual axes,''
  in \emph{2016 IEEE International Conference on Robotics and Automation
  (ICRA)}.\hskip 1em plus 0.5em minus 0.4em\relax IEEE, 2016, pp. 4304--4311.

\bibitem{doer2021yaw}
C.~Doer and G.~F. Trommer, ``Yaw aided radar inertial odometry using manhattan
  world assumptions,'' in \emph{2021 28th Saint Petersburg International
  Conference on Integrated Navigation Systems (ICINS)}.\hskip 1em plus 0.5em
  minus 0.4em\relax IEEE, 2021, pp. 1--9.

\bibitem{geiger2012we}
A.~Geiger, P.~Lenz, and R.~Urtasun, ``Are we ready for autonomous driving? the
  kitti vision benchmark suite,'' in \emph{2012 IEEE conference on computer
  vision and pattern recognition}.\hskip 1em plus 0.5em minus 0.4em\relax IEEE,
  2012, pp. 3354--3361.

\bibitem{6385773}
J.~Sturm, N.~Engelhard, F.~Endres, W.~Burgard, and D.~Cremers, ``A benchmark
  for the evaluation of rgb-d slam systems,'' in \emph{2012 IEEE/RSJ
  International Conference on Intelligent Robots and Systems}, 2012, pp.
  573--580.

\bibitem{lu2020milliego}
C.~X. Lu, M.~R.~U. Saputra, P.~Zhao, Y.~Almalioglu, P.~P. de~Gusmao, C.~Chen,
  K.~Sun, N.~Trigoni, and A.~Markham, ``milliego: single-chip mmwave radar
  aided egomotion estimation via deep sensor fusion,'' in \emph{Proceedings of
  the 18th Conference on Embedded Networked Sensor Systems}, 2020, pp.
  109--122.

\bibitem{rusinkiewicz2001efficient}
S.~Rusinkiewicz and M.~Levoy, ``Efficient variants of the icp algorithm,'' in
  \emph{Proceedings third international conference on 3-D digital imaging and
  modeling}.\hskip 1em plus 0.5em minus 0.4em\relax IEEE, 2001, pp. 145--152.

\bibitem{saputra2020deeptio}
M.~R.~U. Saputra, P.~P. de~Gusmao, C.~X. Lu, Y.~Almalioglu, S.~Rosa, C.~Chen,
  J.~Wahlstr{\"o}m, W.~Wang, A.~Markham, and N.~Trigoni, ``Deeptio: A deep
  thermal-inertial odometry with visual hallucination,'' \emph{IEEE Robotics
  and Automation Letters}, vol.~5, no.~2, pp. 1672--1679, 2020.

\bibitem{saputra2021graph}
M.~R.~U. Saputra, C.~X. Lu, P.~P. de~Gusmao, B.~Wang, A.~Markham, and
  N.~Trigoni, ``Graph-based thermal-inertial slam with probabilistic neural
  networks,'' \emph{arXiv preprint arXiv:2104.07196}, 2021.

\bibitem{clark2017vinet}
R.~Clark, S.~Wang, H.~Wen, A.~Markham, and N.~Trigoni, ``Vinet: Visual-inertial
  odometry as a sequence-to-sequence learning problem,'' in \emph{Proceedings
  of the AAAI Conference on Artificial Intelligence}, vol.~31, no.~1, 2017.

\end{thebibliography}
